\documentclass[preprint]{acmart}

\usepackage{graphicx}
\usepackage{subcaption}

\AtBeginDocument{%
  \providecommand\BibTeX{{%
    \normalfont B\kern-0.5em{\scshape i\kern-0.25em b}\kern-0.8em\TeX}}}

\setcopyright{rightsretained}
\copyrightyear{2025}
\acmYear{2025}

\begin{document}

\title{nanoML for Human Activity Recognition}

\author{Alan T. L. Bacellar}
\email{alanbacellar@utexas.edu}
\affiliation{%
  \institution{The University of Texas at Austin}
  \city{Austin}
  \state{Texas}
  \country{USA}
}

\author{Mugdha P. Jadhao}
\email{mugdhaj17@utexas.edu}
\affiliation{%
  \institution{The University of Texas at Austin}
  \city{Austin}
  \state{Texas}
  \country{USA}
}

\author{Shashank Nag}
\email{shashanknag@utexas.edu}
\affiliation{%
  \institution{The University of Texas at Austin}
  \city{Austin}
  \state{Texas}
  \country{USA}
}

\author{Priscila M. V. Lima}
\email{priscilamvl@cos.ufrj.br}
\affiliation{%
  \institution{Federal University of Rio de Janeiro}
  \city{Rio de Janeiro}
  \state{Rio de Janeiro}
  \country{Brazil}}

\author{Felipe M. G. França}
\email{felipe@ieee.org}
\affiliation{%
  \institution{Instituto de Telecomunicações}
  \city{Porto}
  \state{Porto}
  \country{Portugal}}

\author{Lizy K. John}
\email{ljohn@ece.utexas.edu}
\affiliation{%
  \institution{The University of Texas at Austin}
  \city{Austin}
  \state{Texas}
  \country{USA}}

\renewcommand{\shortauthors}{Bacellar et al.}

\begin{abstract}
  Human Activity Recognition (HAR) is critical for applications in healthcare, fitness, and IoT, but deploying accurate models on resource-constrained devices remains challenging due to high energy and memory demands. This paper demonstrates the application of Differentiable Weightless Neural Networks (DWNs) to HAR, achieving competitive accuracies of \textbf{96.34\%} and \textbf{96.67\%} while consuming only \textbf{56nJ} and \textbf{104nJ} per sample, with an inference time of just \textbf{5ns per sample}. The DWNs were implemented and evaluated on an FPGA, showcasing their pratical feasibility for energy-efficient hardware deployment. DWNs achieve up to \textbf{926,000x energy savings} and \textbf{260x memory reduction} compared to state-of-the-art deep learning methods. These results position DWNs as a nano-machine learning (\textbf{nanoML}) model for HAR, setting a new benchmark in energy efficiency and compactness for edge and wearable devices, paving the way for ultra-efficient edge AI.
\end{abstract}


\maketitle

\section{Introduction}

Human activity recognition (HAR) is a pivotal area of research with applications spanning healthcare, fitness, and workplace monitoring. By enabling the automated identification of activities such as walking, running, or sitting, HAR systems have significantly enhanced our ability to monitor and improve human health and well-being. These advancements are particularly valuable in healthcare, where HAR plays a critical role in early disease detection, rehabilitation monitoring, and elder care. Furthermore, HAR facilitates the development of personalized fitness programs and ensures workplace safety by monitoring posture and movement patterns. With the rapid proliferation of digital health solutions, the demand for accurate, efficient, and scalable HAR systems has grown exponentially.

Wearable devices, such as smartwatches, fitness bands, and IoT-enabled sensors, serve as the primary platforms for HAR. These devices are equipped with sensors like accelerometers and gyroscopes, which continuously collect rich data streams about users' activities. However, the constrained hardware capabilities of wearables pose significant challenges, especially in terms of energy efficiency. As wearables are often battery-powered and expected to operate for extended periods without frequent charging, developing low-energy models for HAR becomes crucial. Energy-efficient models not only extend device battery life but also enable seamless, real-time activity recognition, ensuring their adoption in resource-constrained environments.

Several recent models for HAR have demonstrated impressive accuracy \cite{ignatov, layer_wise, channel_eq_har, harmamba, tslanet}, leveraging advancements in deep learning and other machine learning approaches. These models, however, often rely on architectures with massive computational demands, requiring millions of floating-point operations (FLOPs), or computationally expensive preprocessing steps. Unfortunately, these energy-intensive aspects are frequently overlooked or insufficiently addressed in the presentation of results, creating a disconnect between reported performance and their practical deployment on resource-constrained wearable devices. The mismatch between the energy-efficient hardware of edge devices and the computational needs of these models poses a significant barrier to their widespread adoption in real-world applications.

Recently, Differentiable Weightless Neural Networks (DWNs) have emerged as a promising solution  for edge devices \cite{dwn}, offering energy-efficient models. DWNs are capable of performing inference with minimal energy consumption—often measured in nanojoules (nJ)—and executing in just nanoseconds (ns). While DWNs have demonstrated significant potential in other domains, their application to human activity recognition has not yet been explored.

In this paper, we address the critical gap in energy-efficient HAR solutions by showcasing DWNs as a nano-machine learning approach for human activity recognition (HAR). We demonstrate their ability to achieve accurate recognition with extremely low energy consumption, deploying them on FPGA and highlighting their performance for edge devices. Additionally, we estimate the energy costs of recent state-of-the-art convolutional neural networks (CNNs) \cite{layer_wise, channel_eq_har} and Transformers \cite{harmamba, tslanet}. By comparing these implementations to DWNs, this work not only underscores the potential of DWNs for HAR but also paves the way for further exploration of ultra-efficient models for edge AI applications.

The remainder of the paper is organized as follows - Section \ref{sec:dnn} introduces existing Deep Neural Network based works on Human Activity Recognition, and Section \ref{dwn} introduces DWNs. In Section \ref{sec:exp} we elucidate our experimental methodology and results, and Section \ref{sec:conclusion} concludes the paper.

\begin{figure*}[t]
\centerline{\includegraphics[width=2.0\columnwidth]{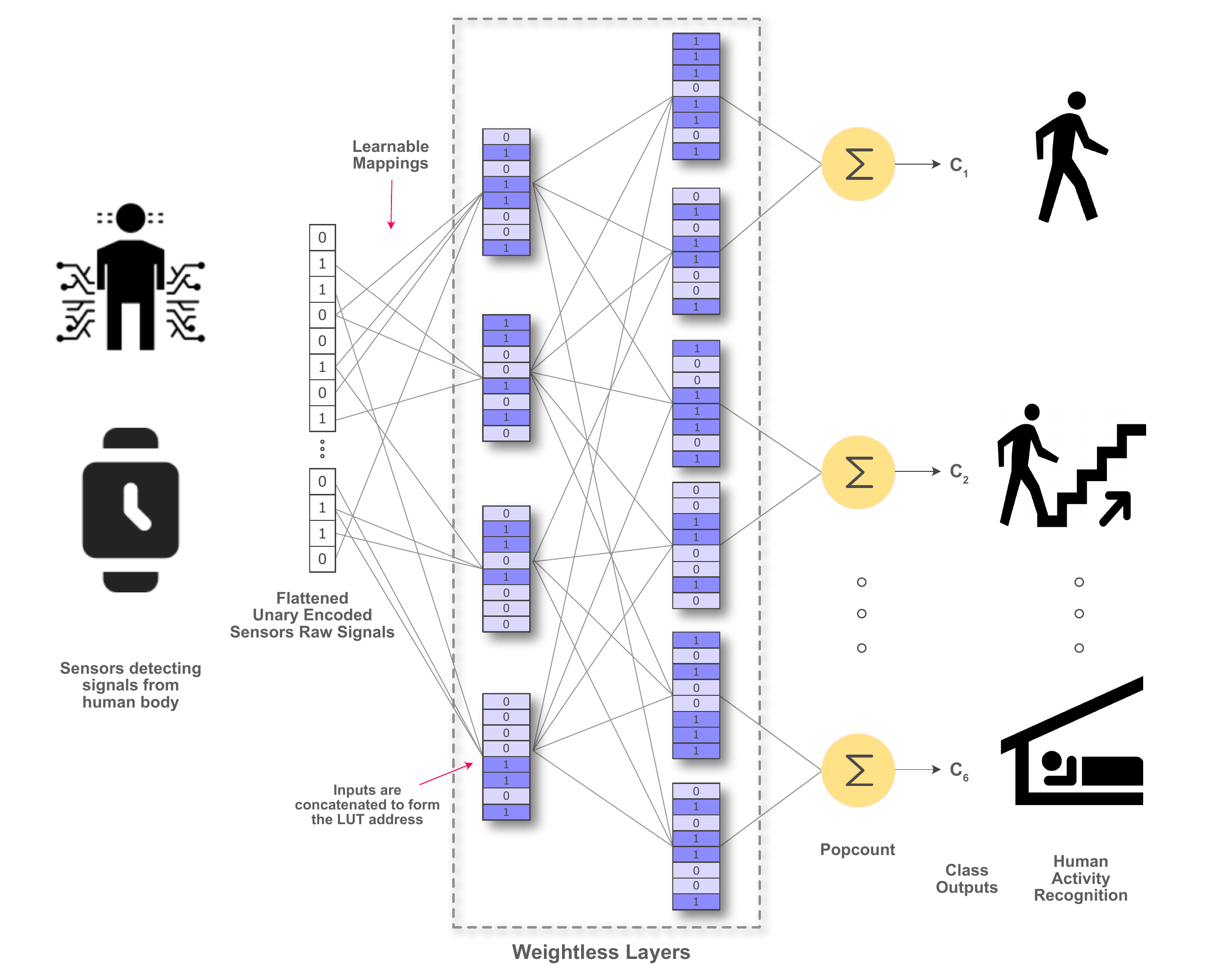}}
\caption{A simple example of a DWN with two layers of LUT-3s. DWNs compute entirely using multiple layers of directly chained lookup tables. Input features are concatenated to form addresses for the first-layer LUTs, whose binary outputs are combined to address the LUTs in subsequent layers. The outputs of the final layer are aggregated to produce activations for each class, without any intermediate arithmetic operations.
}
\label{fig:dwn}
\end{figure*}

\section{Deep Neural Networks Models for Human Activity Recognition}\label{sec:dnn}

Deep neural networks (DNNs) have been extensively explored for human activity recognition (HAR), spannig various architectures, including multi-layer perceptrons (MLPs), Convolutional Neural Networks (CNNs), Transformers, and Hybrid Models. While DNNs have achieved high accuracy on benchmark datasets, including UCI-HAR, their practical deployment on wearable devices remains challenging due to their reliance on computationally expensive preprocessing pipelines and hardware-intensive architectures.

\subsection{Preprocessing in Neural Network-Based HAR}

Many HAR approaches rely on extensive preprocessing steps to extract high-level statistical and spectral features from raw sensor signals. Commonly used features include mean, standard deviation, energy, entropy, and correlation, alongside frequency-domain representations obtained via Fast Fourier Transform (FFT) or wavelet transforms. These preprocessing steps enhance the input data's representation for neural network models, enabling superior accuracy but at a substantial computational cost. For instance, FFT operations require $O(N \log N)$ complexity, which is resource-intensive for edge devices like FPGAs and ASICs. Similarly, the calculation of entropy and correlation involves non-linear operations that are impractical for real-time, low-power environments.

The UCI-HAR dataset~\cite{uci, uci_har_dataset}, a widely used benchmark, exemplifies this dependency on preprocessing. The highest reported accuracy for CNNs on UCI-HAR, 97.62\%, is achieved by elaborate preprocessing pipelines \cite{ignatov}. While effective, such approaches are unsuitable for edge deployment due to their significant preprocessing and computational overhead. 

\subsection{State-of-the-art Models}

Recent studies have demonstrated the feasibility of deploying HAR models directly on raw sensor data without preprocessing. For instance, CNNs applied to raw UCI-HAR data achieved an accuracy of 96.27\%~\cite{layer_wise}. However, this approach incurs a significant computational cost, requiring 33.7 million floating-point operations (FLOPs) per inference. This substantial demand translates to high energy consumption, as quantified in the experimental section of this paper. These findings highlight the trade-off between eliminating preprocessing and managing model complexity, underscoring the need for lightweight, energy-efficient alternatives for resource-constrained environments.

In addition to CNNs, transformers have been explored for HAR due to their ability to model long-term dependencies in sequential data. HARMamba, a transformer-based model, achieved a state-of-the-art accuracy of 97.65\% on UCI-HAR by leveraging self-attention mechanisms \cite{harmamba}. However, the model still relies on some preprocessing, and its large parameter count and complex operations result in significant energy and memory requirements, limiting its suitability for edge devices. Similarly, TSLANet, another transformer-based architecture, tailored for time series, achieved 96.06\% accuracy \cite{tslanet}. Despite their strong performance, transformer models remain computationally intensive, necessitating millions of FLOPs per inference. These challenges underscore the demand for more energy-efficient architectures to meet the constraints of edge and wearable devices.

\section{Differentiable Weightless Neural Networks (DWN)}\label{dwn}

Weightless Neural Networks (WNNs) \cite{wnn_intro_esann} are a class of neural architectures that uses lookup tables (LUTs) as neurons instead of weighted connections and dot products used by DNNs. The expressive power of a neural network neuron is characterized by its Vapnik-Chervonenkis (VC) dimension, which quantifies the model's capacity to distinguish complex patterns. Notably, the VC dimension of a LUT scales exponentially with the number of inputs $n$, as $2^n$ \cite{vc_wi}, whereas, for a conventional deep neural network (DNN) neuron with $n$ inputs, the VC dimension is limited to $n+1$. For example, to accurately model a simple XOR pattern of 2 bits, a WNN requires only one LUT with 4 entries. In contrast, a DNN would need 3 neurons, resulting in 6 weights and 3 biases, totalizing 9 parameters. This stark difference highlights the efficiency of LUT-based models in representing complex patterns within a single computational unit compared to the relatively parameter-intensive approach of DNNs.

Despite their theoretical advantages, traditional WNNs have primarily been confined to single-layer architectures and random connections of inputs to LUTs. This limitation arises from the non-differentiable nature of LUTs, which has historically posed challenges for gradient-based optimization techniques, impeding the development of multi-layer architectures.

Recently, Differentiable Weightless Neural Networks (DWNs) \cite{dwn} have been proposed to address this limitation by introducing differentiability to LUT-based architectures, thereby enabling the training of multi-layer WNNs. These networks achieve up to a 286× reduction in energy costs and a 60× reduction in latency compared to FINN \cite{finn}, the state-of-the-art Binary Neural Network (BNN) framework, under iso-accuracy scenarios.

 \subsection{DWN Architecture and Training}

DWN architecture is built entirely from LUTs, routing connections, and a final population count (popcount) operation.

As illustrated in Figure~\ref{fig:dwn}, DWNs operate on binary inputs, which are routed to form the addresses for the first layer of LUTs. Each LUT indexes trainable binary values corresponding to these addresses, producing binary outputs. These outputs serve as inputs for the next layer, propagating through the network without any intermediate arithmetic operations. The outputs from the final layer of LUTs are aggregated using a popcount operation, which calculates the total number of binary "1s" across the outputs. This aggregated value determines the activations for each class.

To address the challenge of training discrete LUT-based networks, DWNs propose the Extended Finite Difference (EFD) technique. EFD introduces differentiability to LUTs by approximating gradients for the discrete entries and indexing operations within the tables. This enables LUT values to be optimized through gradient-based learning, allowing DWNs to be trained end-to-end like traditional neural networks.

Another critical innovation of DWNs is the concept of Learnable Mapping, which enables the connections between inputs and LUTs, and between layers of LUTs, to be learned during training. Importantly, these learned connections are fixed at inference time, introducing no additional overhead as routing becomes a static design. This ensures that DWNs remain computationally efficient during deployment.

The minimalistic design of DWNs—comprising LUTs, routing, and a final popcount operation—makes them highly suitable for hardware implementation. On FPGAs, LUTs directly map to hardware primitives, and routing leverages the programmable interconnects of the device. On ASICs, the simplicity of DWNs ensures a compact design with minimal area and power requirements. These characteristics enable DWNs to achieve real-time inference in energy-constrained environments, such as wearable devices and IoT applications.

\subsection{Thermometer Unary Encoding}

Weightless Neural Networks (WNNs) require binary inputs, making the method of binarizing continuous sensor data critical for performance \cite{binary_encodings}. Among the available strategies, thermometer encoding is the state-of-the-art approach.

Thermometer encoding \cite{thermometer, distributive} represents numerical values as unary binary vectors, where the number of consecutive "1s" corresponds to the magnitude of the value. For instance, a reading of 3 in a range of 5 would be encoded as `[1, 1, 1, 0, 0]`.

Importantly, thermometer encoding is not part of the DWN model itself and can be implemented directly at the sensor level. This allows the sensor to output thermometer-encoded binary values, avoiding additional binarization steps. Its simplicity and lightweight implementation make it an ideal choice for deployment on resource-constrained hardware. Furthermore, sensors designed to output thermometer-encoded values need only recognize a small, discrete set of values rather than full precision (e.g., FP32), significantly reducing sensor complexity and cost.

\section{Experimental Evaluation}\label{sec:exp}

\begin{table*}[]
\caption{Comparison of state-of-the-art models for Human Activity Recognition (HAR) on the UCI-HAR dataset on the Xilinx XC7Z020CLG400 FPGA. DWNs achieve significant energy efficiency compared to CNN and Transformer-based architectures.}
\label{tab:results1}
\begin{tabular}{@{}lccccc@{}}
\toprule
\textbf{Model}           & \textbf{Accuracy} & \textbf{F1 Score} & \textbf{\begin{tabular}[c]{@{}c@{}}Model Size\\ (KiB)\end{tabular}} & \textbf{FLOPs} & \textbf{Energy/Sample} \\ \midrule
TSLANet \cite{tslanet}                 & 96.06\%             & -           & -                                                                   & 69M            & 52mJ                   \\
Channel-Equalization-HAR \cite{channel_eq_har} & 97.35\%             & 97.12\%       & 1600                                                                & 44M         & 33mJ                   \\
CNN \cite{layer_wise}         & 96.27\%             & 96.27\%       & 5100                                                                & 35M         & 26mJ                   \\
HARMamba \cite{harmamba}                & 97.65\%             & 97.01\%       & 1300                                                                & 11M         & 8mJ                    \\
DWN                    & 96.34\%             & 96.30\%       & 19                                                                  & 0              & 0.000056mJ             \\
DWN                     & 96.67\%             & 96.68\%       & 39                                                                  & 0              & 0.000104mJ             \\ \bottomrule
\end{tabular}
\end{table*}


\subsection{Dataset Description}

The \textbf{UCI Human Activity Recognition (HAR) Dataset} \cite{uci_har_dataset} is a widely used benchmark for evaluating human activity recognition models. The dataset was collected from 30 participants aged between 19 and 48 years as they performed six basic activities: \textit{walking}, \textit{walking upstairs}, \textit{walking downstairs}, \textit{sitting}, \textit{standing}, and \textit{lying down}. Each participant carried a smartphone (Samsung Galaxy S II) on their waist, which recorded sensor data from a 3-axis accelerometer and a 3-axis gyroscope.

\paragraph{Raw Signals}
The raw sensor data consists of 128 samples per activity segment, corresponding to a window of 2.56 seconds sampled at 50 Hz. These raw signals capture the accelerometer and gyroscope readings along three spatial axes (X, Y, Z), providing a direct representation of the participants' movements.

\paragraph{Hand-Crafted Features}
In addition to the raw signals, the dataset includes 561 hand-crafted features derived from the time and frequency domains. These features, used by many traditional models, include statistical metrics such as mean, standard deviation, skewness, and kurtosis, as well as frequency-domain features obtained via the Fast Fourier Transform (FFT). While these features enhance the representational power of the data, their extraction is computationally and energetically expensive, posing significant challenges for deployment on hardware-constrained wearable devices. 

In this work, we deploy our model exclusively on the raw 128-sample signals, excluding the hand-crafted signals and without using any preprocessing on the signals. By eliminating the need for feature extraction, we demonstrate the feasibility of achieving accurate human activity recognition while adhering to the strict energy constraints of real-world wearable devices.

\paragraph{Data Splits}
The dataset is pre-divided into training and testing sets in an inter-patient fashion, where data from 70\% of the participants are used for training, and the remaining 30\% are reserved for testing. This ensures that the model is evaluated on entirely unseen participants, providing a robust assessment of its ability to generalize to new users.


\subsection{Thermometer Encoding}
Each of the 9 raw sensor signals in the dataset is converted to unary encoding using a 20-bit Distributive Thermometer Encoding scheme \cite{distributive}, as described in section 3.2.

\subsection{Model Configurations and Training Details}

To identify the optimal DWN configuration, we split the training set into a training subset (80\% of the training data) and a validation subset (20\%) for hyperparameter search and Neural Architecture Search (NAS). The search explored the following ranges:
\begin{itemize}
    \item \textbf{Number of Layers}: [1, 2, 3, 4]
    \item \textbf{LUT Size}: [2, 3, 4, 5, 6, 7, 8]
    \item \textbf{Softmax Temperature} (\(\tau\)): \(\left[\frac{1}{0.5}, \frac{1}{0.3}, \frac{1}{0.1}, \frac{1}{0.05}, \frac{1}{0.03}, \frac{1}{0.01}\right]\)

\end{itemize}

The best-performing architecture consists of a single layer of LUT-4 with a softmax temperature of \(\tau = \frac{1}{0.03}\).

\subsubsection{Data Augmentation}
To enhance generalization, we applied a range of standard 1D signal augmentations to the training data. Each augmentation was applied with a probability of \(p = 0.3\), including:
\begin{itemize}
    \item \textbf{Time Shift}: Randomly shifts signals along the time axis by up to 10 steps.
    \item \textbf{Scaling}: Multiplies signals by a random scaling factor uniformly sampled from [0.9, 1.1].
    \item \textbf{Jitter}: Adds Gaussian noise with a standard deviation of 0.05 to simulate signal variability.
    \item \textbf{Time Masking}: Masks a random section of the signal (up to 10 time steps) by setting it to zero.
    \item \textbf{Axis Flip}: Randomly inverts one or more signal axes.
    \item \textbf{Rotation}: Applies a small rotation (up to \(10^\circ\)) to the first three axes, simulating spatial transformations.
    \item \textbf{Low-Pass Filtering}: Smoothens the signal using a low-pass Butterworth filter with a cutoff frequency of 20 Hz.
\end{itemize}

\subsubsection{Training Procedure}
The DWN was trained using the Adam optimizer with the following settings:
\begin{itemize}
    \item \textbf{Batch Size}: 100
    \item \textbf{Learning Rate}: 0.01, decayed by a factor of 0.1 every 14 epochs.
    \item \textbf{Total Training Epochs}: 32
\end{itemize}

We trained two DWN models for evaluation: one with 10,000 LUTs and another with 20,000 LUTs. The training process was conducted using the full training set, and the final models were obtained after completing all 32 epochs. This ensures the model fully leverages the available training data to achieve optimal performance.

\subsection{FPGA Board Deployment}

We deployed the DWN models on the Xilinx XC7Z020CLG400 FPGA, operating in out-of-context mode. This configuration allows us to isolate and demonstrate the model's computational capabilities independently of specific data transfer constraints. We write SystemVerilog RTL codes for the design using mako templating scripts \cite{mako} based on the hardware architecture proposed by the authors of DWN \cite{dwn}. The design is synthesized on the target FPGA using the Xilinx Vivado 2022.1 tool, with a target clock of 200 MHz. This enables us to determine the maximum achievable frequency for the design by observing the reported Worst Negative Slack (WNS). The DWN architecture, implemented with pipelining, achieves one inference per clock cycle, ensuring maximal throughput under ideal conditions. Any bottlenecks in real-world deployment would arise solely from data transfer rates, not the model's computation.

We evaluate the energy consumption for DWN models by performing a vectorless power estimation of the synthesized design on our target FPGA with a default switching activity of 12.5\%. We report the energy per sample inference of these models as the energy spent in operating the deployment for an amount of time corresponding to the latency of execution. 
\subsection{Energy Estimation for Other Models}

To compare DWN's energy efficiency with other models, we estimated the energy consumption of prior work based on the FLOP counts reported in their respective papers. We consider Verilog-based RTL designs for individual floating-point operation units and synthesize these on the same target Xilinx XC7Z020CLG400 FPGA with Xilinx Vivado 2022.1 to report vectorless power estimation for a 12.5\% switching activity. With this approach, the energy per floating-point operation (FLOP) on the FPGA was determined as:
\begin{itemize}
    \item \textbf{Multiplication (FP32)}: 0.928nJ
    \item \textbf{Addition (FP32)}: 0.594nJ
\end{itemize}

On average, a FLOP in neural network architectures involves roughly equal proportions of multiplications and additions, as a DNN neuron with \(n\) inputs performs \(n\) multiplications and \(n\) additions. Using this assumption, we estimated the energy per FLOP as the average of the two operations: 0.761\text{nJ}.

The total energy consumption of other models was calculated by multiplying the reported FLOP counts from their respective papers with this estimated energy per FLOP. It is important to note that we do not know whether the reported FLOP counts include only the model computations or also account for any preprocessing steps required for those architectures. If preprocessing FLOPs are excluded, it puts these models at an advantage over DWNs in our comparison.

\subsection{Results}

\begin{table*}[]
\caption{Detailed performance of DWNs for HAR on the Xilinx XC7Z020CLG400 FPGA deployed in out-of-context mode. The 'Time/Sample' metric is calculated as the reciprocal of throughput, and 'Latency' represents the end-to-end processing time of the model.}
\label{tab:results2}
\begin{tabular}{@{}lcccccccc@{}}
\toprule
\textbf{Model} & \textbf{Accuracy} & \textbf{Model Size} & \textbf{\#LUTs} & \textbf{\#FFs} & \textbf{FMax} & \textbf{Latency} & \textbf{Time/Sample} & \textbf{Energy/Sample} \\ \midrule
DWN            & 96.34\%             & 19.5KiB             & 20444           & 13948          & 199MHz        & 40ns             & 5ns                  & 56nJ                   \\
DWN            & 96.67\%             & 39.1KiB             & 41252           & 27861          & 199MHz        & 45ns             & 5ns                  & 104nJ                  \\ \bottomrule
\end{tabular}
\end{table*}

The application of Differentiable Weightless Neural Networks (DWNs) to Human Activity Recognition (HAR) underscores their unparalleled energy efficiency and suitability for edge deployment compared to state-of-the-art methods.

\paragraph{Accuracy and Energy Efficiency}
The DWN models achieve accuracies of \textbf{96.34\%} and \textbf{96.67\%}, closing the gap with the state-of-the-art HAR methods while consuming just \textbf{56nJ} and \textbf{104nJ} per sample. These energy consumption levels represent a dramatic \textbf{77,000x to 926,000x reduction} compared to other models, which require between 8mJ and 52mJ per inference.

\begin{figure}[ht]
    \centering
    \begin{subfigure}[b]{0.45\textwidth}
        \centering
        \includegraphics[width=\textwidth]{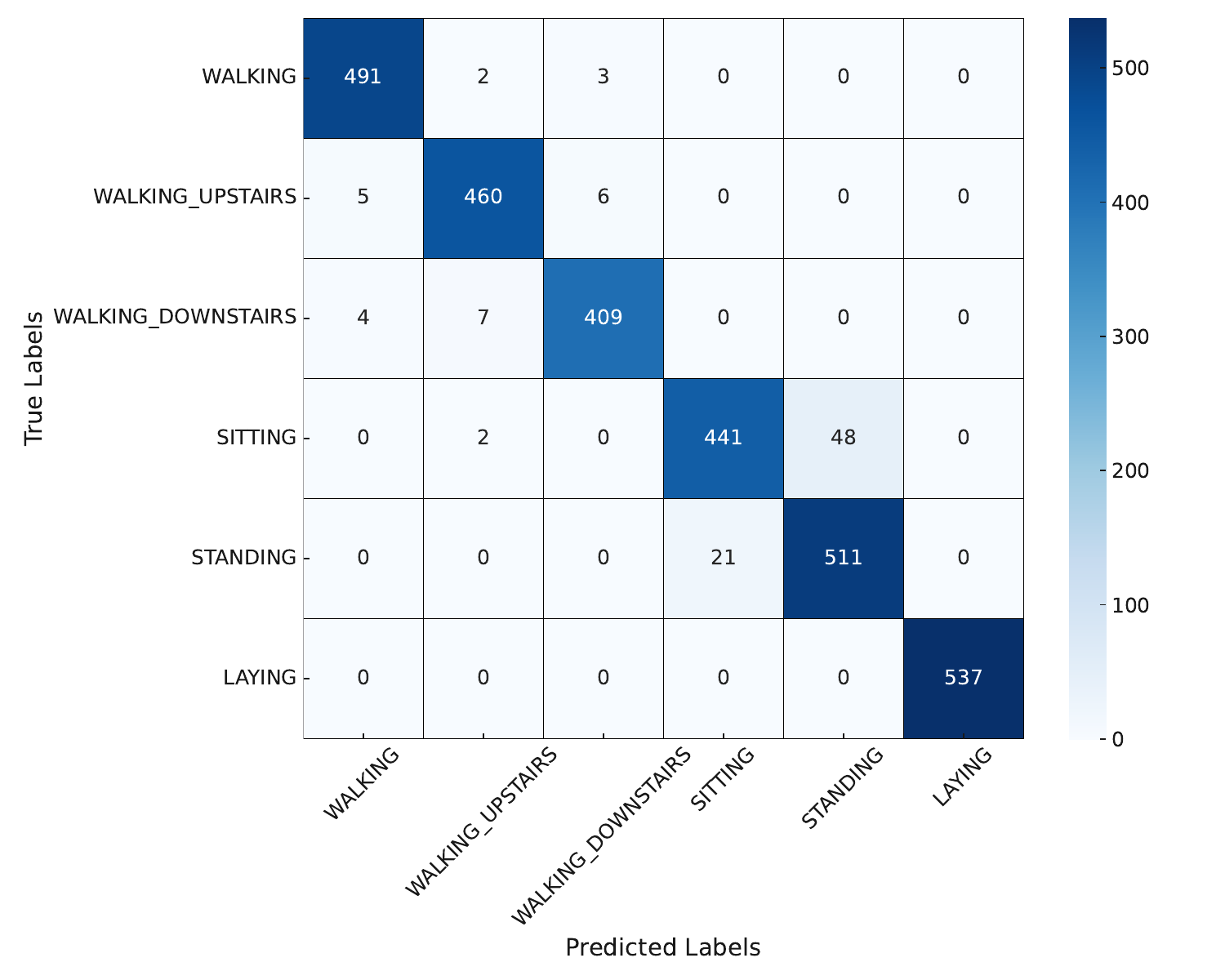}
        \caption{Confusion matrix for the DWN model with a F1 of 96.30\%}
        \label{fig:original_matrix}
    \end{subfigure}
    \hfill
    \begin{subfigure}[b]{0.45\textwidth}
        \centering
        \includegraphics[width=\textwidth]{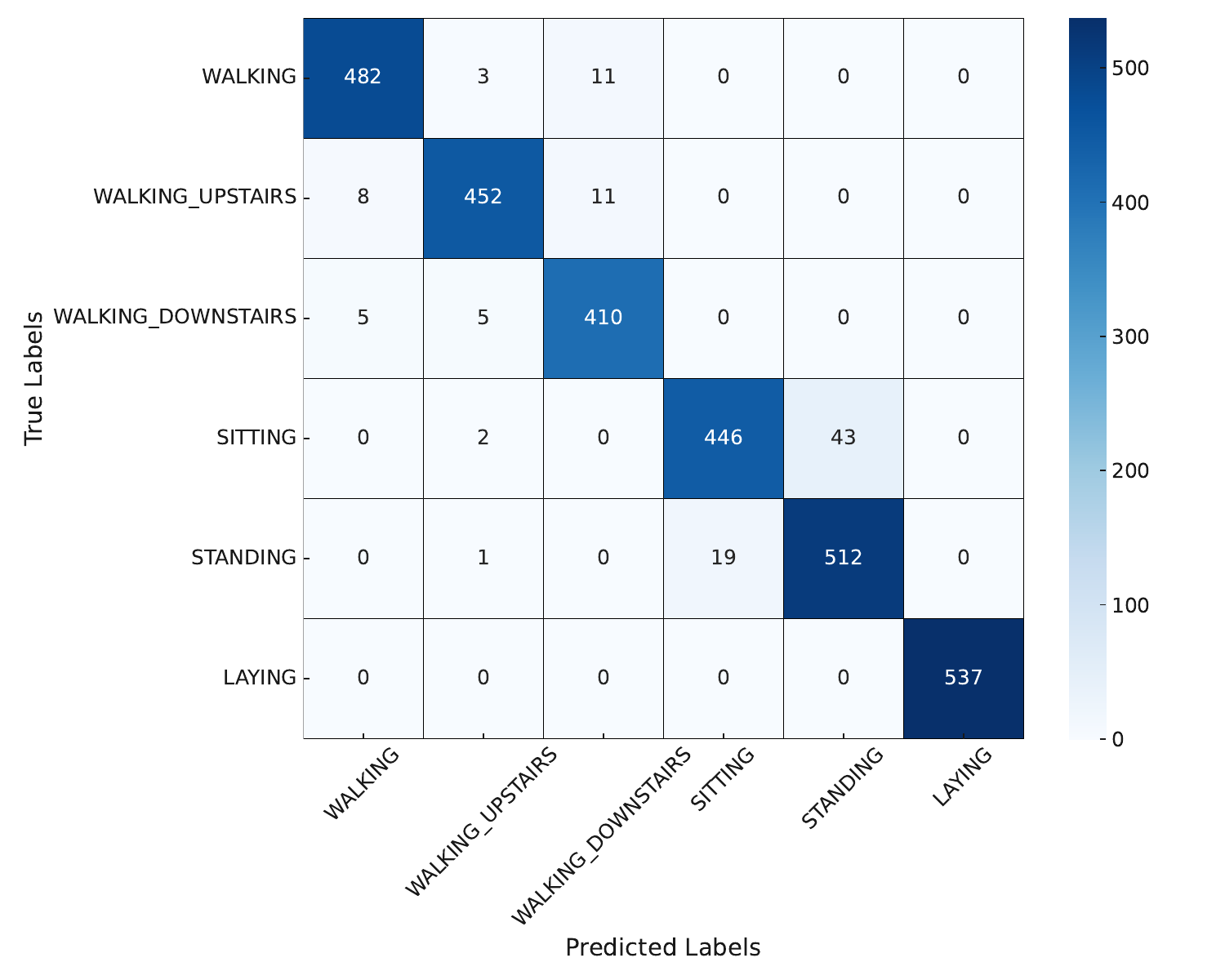}
        \caption{Confusion matrix for the DWN model with a F1 of 96.68\%}
        \label{fig:updated_matrix}
    \end{subfigure}
    \caption{Confusion matrices for Differentiable Weightless Neural Network (DWN) models evaluated on the UCI-HAR dataset.}
    \label{fig:side_by_side_matrices}
\end{figure}

\paragraph{Model Size and Compactness} The DWN models are significantly smaller than traditional state-of-the-art solutions. With sizes of \textbf{19.5KiB} and \textbf{39.1KiB}, they achieve a \textbf{33x to 260x reduction} compared to models such as CNN and HARMamba, which range from 1.3MiB to 5.1MiB. This compactness, combined with their hardware-friendly design, supports their deployment on resource-constrained devices.

\paragraph{Hardware Performance}
Unlike traditional neural networks, DWNs operate without any floating-point operations (FLOPs). Instead, they rely entirely on LUTs, routing, and a final population count (popcount) operation, which drastically reduces computation overhead. The DWN models perform \textbf{one inference per clock cycle}, resulting in a total processing time of \textbf{5ns per sample} on the Xilinx XC7Z020CLG400 FPGA, running at \textbf{199MHz}. This fully pipelined design ensures consistently low inference times, further emphasizing their potential for real-time HAR applications.

\section{Conclusion and Future Work}\label{sec:conclusion}

This study demonstrates that Differentiable Weightless Neural Networks (DWNs) embody the principles of nano-machine learning (nanoML), achieving unprecedented levels of energy efficiency and compactness in Human Activity Recognition (HAR). DWNs deliver competitive accuracies of 96.34\% and 96.67\% while consuming just 56nJ and 104nJ per sample, setting a new benchmark for energy-efficient HAR systems.

In comparison to state-of-the-art models, DWNs achieve up to 926,000x energy savings and a 260x reduction in memory size, while maintaining low latency with a processing time of 5ns per sample. Furthermore, DWNs can be directly converted into logic gates \cite{dwn} and implemented as custom ASICs, making them ready for immediate integration into wearable devices.

Future work will investigate the extension of DWNs to multi-modal sensor data and explore their scalability for more complex HAR scenarios. Additionally, efforts will focus on designing custom ASICs for DWNs to further enhance their efficiency in wearable applications, optimizing DWNs for diverse hardware platforms, and expanding their applications beyond HAR.

\begin{acks}
This research was supported in part by the Semiconductor Research Corporation (SRC) Task 3148.001, NSF Grants \#2326894, \#2425655, and NVIDIA Applied Research Accelerator Program Grant.
\end{acks}

\bibliographystyle{ACM-Reference-Format}
\bibliography{bibliography}


\begin{thebibliography}{15}


\ifx \showCODEN    \undefined \def \showCODEN     #1{\unskip}     \fi
\ifx \showDOI      \undefined \def \showDOI       #1{#1}\fi
\ifx \showISBNx    \undefined \def \showISBNx     #1{\unskip}     \fi
\ifx \showISBNxiii \undefined \def \showISBNxiii  #1{\unskip}     \fi
\ifx \showISSN     \undefined \def \showISSN      #1{\unskip}     \fi
\ifx \showLCCN     \undefined \def \showLCCN      #1{\unskip}     \fi
\ifx \shownote     \undefined \def \shownote      #1{#1}          \fi
\ifx \showarticletitle \undefined \def \showarticletitle #1{#1}   \fi
\ifx \showURL      \undefined \def \showURL       {\relax}        \fi
\providecommand\bibfield[2]{#2}
\providecommand\bibinfo[2]{#2}
\providecommand\natexlab[1]{#1}
\providecommand\showeprint[2][]{arXiv:#2}

\bibitem[Aleksander et~al\mbox{.}(2009)]%
        {wnn_intro_esann}
\bibfield{author}{\bibinfo{person}{Igor Aleksander}, \bibinfo{person}{Massimo
  De~Gregorio}, \bibinfo{person}{Felipe França}, \bibinfo{person}{Priscila
  Lima}, {and} \bibinfo{person}{Helen Morton}.}
  \bibinfo{year}{2009}\natexlab{}.
\newblock \showarticletitle{A brief introduction to Weightless Neural Systems}.
  In \bibinfo{booktitle}{\emph{17th European Symposium on Artificial Neural
  Networks (ESANN)}}. \bibinfo{pages}{299--305}.
\newblock


\bibitem[Anguita et~al\mbox{.}(2013)]%
        {uci_har_dataset}
\bibfield{author}{\bibinfo{person}{D. Anguita}, \bibinfo{person}{Alessandro
  Ghio}, \bibinfo{person}{L. Oneto}, \bibinfo{person}{Xavier Parra}, {and}
  \bibinfo{person}{Jorge~Luis Reyes-Ortiz}.} \bibinfo{year}{2013}\natexlab{}.
\newblock \showarticletitle{A Public Domain Dataset for Human Activity
  Recognition using Smartphones}. In \bibinfo{booktitle}{\emph{The European
  Symposium on Artificial Neural Networks}}.
\newblock
\urldef\tempurl%
\url{https://api.semanticscholar.org/CorpusID:6975432}
\showURL{%
\tempurl}


\bibitem[Bacellar et~al\mbox{.}(2022)]%
        {distributive}
\bibfield{author}{\bibinfo{person}{Alan Bacellar}, \bibinfo{person}{Zachary
  Susskind}, \bibinfo{person}{Luis Villon}, \bibinfo{person}{Igor Miranda},
  \bibinfo{person}{Leandro Santiago}, \bibinfo{person}{Diego Dutra},
  \bibinfo{person}{Mauricio Jr}, \bibinfo{person}{LIZY JOHN},
  \bibinfo{person}{Priscila Lima}, {and} \bibinfo{person}{Felipe França}.}
  \bibinfo{year}{2022}\natexlab{}.
\newblock \showarticletitle{Distributive Thermometer: A New Unary Encoding for
  Weightless Neural Networks}. \bibinfo{pages}{31--36}.
\newblock
\urldef\tempurl%
\url{https://doi.org/10.14428/esann/2022.ES2022-94}
\showDOI{\tempurl}


\bibitem[Bacellar et~al\mbox{.}(2024)]%
        {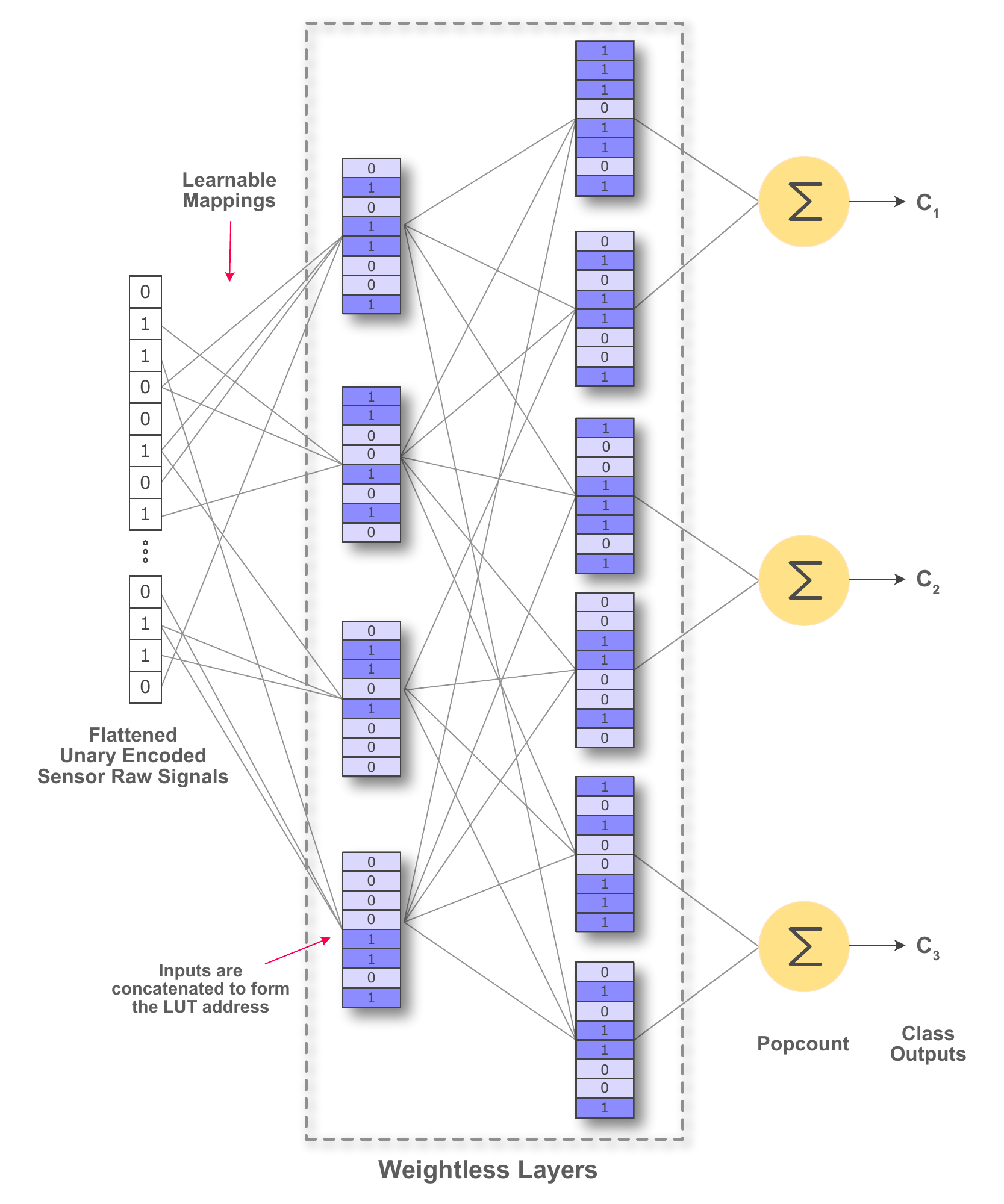}
\bibfield{author}{\bibinfo{person}{Alan T.~L. Bacellar},
  \bibinfo{person}{Zachary Susskind}, \bibinfo{person}{Mauricio Breternitz~Jr},
  \bibinfo{person}{Eugene John}, \bibinfo{person}{Lizy~Kurian John},
  \bibinfo{person}{Priscila Machado~Vieira Lima}, {and}
  \bibinfo{person}{Felipe~M.G. Fran\c{c}a}.} \bibinfo{year}{2024}\natexlab{}.
\newblock \showarticletitle{Differentiable Weightless Neural Networks}. In
  \bibinfo{booktitle}{\emph{Proceedings of the 41st International Conference on
  Machine Learning}} \emph{(\bibinfo{series}{Proceedings of Machine Learning
  Research}, Vol.~\bibinfo{volume}{235})},
  \bibfield{editor}{\bibinfo{person}{Ruslan Salakhutdinov},
  \bibinfo{person}{Zico Kolter}, \bibinfo{person}{Katherine Heller},
  \bibinfo{person}{Adrian Weller}, \bibinfo{person}{Nuria Oliver},
  \bibinfo{person}{Jonathan Scarlett}, {and} \bibinfo{person}{Felix
  Berkenkamp}} (Eds.). \bibinfo{publisher}{PMLR}, \bibinfo{pages}{2277--2295}.
\newblock
\urldef\tempurl%
\url{https://proceedings.mlr.press/v235/bacellar24a.html}
\showURL{%
\tempurl}


\bibitem[Bayer(2021)]%
        {mako}
\bibfield{author}{\bibinfo{person}{Michael Bayer}.}
  \bibinfo{year}{2021}\natexlab{}.
\newblock \bibinfo{title}{Mako Templates for Python}.
\newblock
\newblock
\urldef\tempurl%
\url{https://www.makotemplates.org/}
\showURL{%
\tempurl}


\bibitem[Carneiro et~al\mbox{.}(2015)]%
        {thermometer}
\bibfield{author}{\bibinfo{person}{Hugo Carneiro}, \bibinfo{person}{Felipe
  França}, {and} \bibinfo{person}{Priscila Lima}.}
  \bibinfo{year}{2015}\natexlab{}.
\newblock \showarticletitle{Multilingual part-of-speech tagging with weightless
  neural networks}.
\newblock \bibinfo{journal}{\emph{Neural Networks}}  \bibinfo{volume}{66}
  (\bibinfo{date}{03} \bibinfo{year}{2015}).
\newblock
\urldef\tempurl%
\url{https://doi.org/10.1016/j.neunet.2015.02.012}
\showDOI{\tempurl}


\bibitem[Carneiro et~al\mbox{.}(2019)]%
        {vc_wi}
\bibfield{author}{\bibinfo{person}{Hugo C.~C. Carneiro},
  \bibinfo{person}{Carlos~Eduardo Pedreira}, \bibinfo{person}{Felipe
  Maia~Galv{\~a}o França}, {and} \bibinfo{person}{Priscila Machado~Vieira
  Lima}.} \bibinfo{year}{2019}\natexlab{}.
\newblock \showarticletitle{The Exact {VC} Dimension of the {WiSARD} n-Tuple
  Classifier}.
\newblock \bibinfo{journal}{\emph{Neural Computation}}  \bibinfo{volume}{31}
  (\bibinfo{year}{2019}), \bibinfo{pages}{176--207}.
\newblock
\urldef\tempurl%
\url{https://api.semanticscholar.org/CorpusID:53715711}
\showURL{%
\tempurl}


\bibitem[Dua and Graff(2017)]%
        {uci}
\bibfield{author}{\bibinfo{person}{Dheeru Dua} {and} \bibinfo{person}{Casey
  Graff}.} \bibinfo{year}{2017}\natexlab{}.
\newblock \bibinfo{title}{{UCI} Machine Learning Repository}.
\newblock
\newblock
\urldef\tempurl%
\url{http://archive.ics.uci.edu/ml}
\showURL{%
\tempurl}


\bibitem[Eldele et~al\mbox{.}(2024)]%
        {tslanet}
\bibfield{author}{\bibinfo{person}{Emadeldeen Eldele}, \bibinfo{person}{Mohamed
  Ragab}, \bibinfo{person}{Zhenghua Chen}, \bibinfo{person}{Min Wu}, {and}
  \bibinfo{person}{Xiaoli Li}.} \bibinfo{year}{2024}\natexlab{}.
\newblock \showarticletitle{{TSLAN}et: Rethinking Transformers for Time Series
  Representation Learning}. In \bibinfo{booktitle}{\emph{Proceedings of the
  41st International Conference on Machine Learning}}
  \emph{(\bibinfo{series}{Proceedings of Machine Learning Research},
  Vol.~\bibinfo{volume}{235})}, \bibfield{editor}{\bibinfo{person}{Ruslan
  Salakhutdinov}, \bibinfo{person}{Zico Kolter}, \bibinfo{person}{Katherine
  Heller}, \bibinfo{person}{Adrian Weller}, \bibinfo{person}{Nuria Oliver},
  \bibinfo{person}{Jonathan Scarlett}, {and} \bibinfo{person}{Felix
  Berkenkamp}} (Eds.). \bibinfo{publisher}{PMLR},
  \bibinfo{pages}{12409--12428}.
\newblock
\urldef\tempurl%
\url{https://proceedings.mlr.press/v235/eldele24a.html}
\showURL{%
\tempurl}


\bibitem[Huang et~al\mbox{.}(2023)]%
        {channel_eq_har}
\bibfield{author}{\bibinfo{person}{Wenbo Huang}, \bibinfo{person}{Lei Zhang},
  \bibinfo{person}{Hao Wu}, \bibinfo{person}{Fuhong Min}, {and}
  \bibinfo{person}{Aiguo Song}.} \bibinfo{year}{2023}\natexlab{}.
\newblock \showarticletitle{Channel-Equalization-HAR: A Light-weight
  Convolutional Neural Network for Wearable Sensor Based Human Activity
  Recognition}.
\newblock \bibinfo{journal}{\emph{IEEE Transactions on Mobile Computing}}
  \bibinfo{volume}{22}, \bibinfo{number}{9} (\bibinfo{year}{2023}),
  \bibinfo{pages}{5064--5077}.
\newblock
\urldef\tempurl%
\url{https://doi.org/10.1109/TMC.2022.3174816}
\showDOI{\tempurl}


\bibitem[Ignatov(2018)]%
        {ignatov}
\bibfield{author}{\bibinfo{person}{Andrey Ignatov}.}
  \bibinfo{year}{2018}\natexlab{}.
\newblock \showarticletitle{Real-time human activity recognition from
  accelerometer data using Convolutional Neural Networks}.
\newblock \bibinfo{journal}{\emph{Applied Soft Computing}}
  \bibinfo{volume}{62} (\bibinfo{year}{2018}), \bibinfo{pages}{915--922}.
\newblock
\showISSN{1568-4946}
\urldef\tempurl%
\url{https://doi.org/10.1016/j.asoc.2017.09.027}
\showDOI{\tempurl}


\bibitem[Kappaun et~al\mbox{.}(2016)]%
        {binary_encodings}
\bibfield{author}{\bibinfo{person}{Andressa Kappaun}, \bibinfo{person}{Karine
  Camargo}, \bibinfo{person}{Fabio Rangel}, \bibinfo{person}{Fabricio Faria},
  \bibinfo{person}{Priscila Lima}, {and} \bibinfo{person}{Jonice Oliveira}.}
  \bibinfo{year}{2016}\natexlab{}.
\newblock \showarticletitle{Evaluating Binary Encoding Techniques for
  {{WiSARD}}}. In \bibinfo{booktitle}{\emph{BRACIS}}.
  \bibinfo{pages}{103--108}.
\newblock
\urldef\tempurl%
\url{https://doi.org/10.1109/BRACIS.2016.029}
\showDOI{\tempurl}


\bibitem[Li et~al\mbox{.}(2024)]%
        {harmamba}
\bibfield{author}{\bibinfo{person}{Shuangjian Li}, \bibinfo{person}{Tao Zhu},
  \bibinfo{person}{Furong Duan}, \bibinfo{person}{Liming Chen},
  \bibinfo{person}{Huansheng Ning}, \bibinfo{person}{Christopher Nugent}, {and}
  \bibinfo{person}{Yaping Wan}.} \bibinfo{year}{2024}\natexlab{}.
\newblock \showarticletitle{HARMamba: Efficient and Lightweight Wearable Sensor
  Human Activity Recognition Based on Bidirectional Mamba}.
\newblock \bibinfo{journal}{\emph{IEEE Internet of Things Journal}}
  (\bibinfo{year}{2024}), \bibinfo{pages}{1--1}.
\newblock
\urldef\tempurl%
\url{https://doi.org/10.1109/JIOT.2024.3463405}
\showDOI{\tempurl}


\bibitem[Tang et~al\mbox{.}(2021)]%
        {layer_wise}
\bibfield{author}{\bibinfo{person}{Yin Tang}, \bibinfo{person}{Qi Teng},
  \bibinfo{person}{Lei Zhang}, \bibinfo{person}{Fuhong Min}, {and}
  \bibinfo{person}{Jun He}.} \bibinfo{year}{2021}\natexlab{}.
\newblock \showarticletitle{Layer-Wise Training Convolutional Neural Networks
  With Smaller Filters for Human Activity Recognition Using Wearable Sensors}.
\newblock \bibinfo{journal}{\emph{IEEE Sensors Journal}} \bibinfo{volume}{21},
  \bibinfo{number}{1} (\bibinfo{year}{2021}), \bibinfo{pages}{581--592}.
\newblock
\urldef\tempurl%
\url{https://doi.org/10.1109/JSEN.2020.3015521}
\showDOI{\tempurl}


\bibitem[Umuroglu et~al\mbox{.}(2017)]%
        {finn}
\bibfield{author}{\bibinfo{person}{Yaman Umuroglu},
  \bibinfo{person}{Nicholas~J. Fraser}, \bibinfo{person}{Giulio Gambardella},
  \bibinfo{person}{Michaela Blott}, \bibinfo{person}{Philip Leong},
  \bibinfo{person}{Magnus Jahre}, {and} \bibinfo{person}{Kees Vissers}.}
  \bibinfo{year}{2017}\natexlab{}.
\newblock \showarticletitle{FINN: A Framework for Fast, Scalable Binarized
  Neural Network Inference}. In \bibinfo{booktitle}{\emph{Proceedings of the
  2017 ACM/SIGDA International Symposium on Field-Programmable Gate Arrays}}
  (Monterey, California, USA) \emph{(\bibinfo{series}{FPGA '17})}.
  \bibinfo{publisher}{Association for Computing Machinery},
  \bibinfo{address}{New York, NY, USA}, \bibinfo{pages}{65–74}.
\newblock
\showISBNx{9781450343541}
\urldef\tempurl%
\url{https://doi.org/10.1145/3020078.3021744}
\showDOI{\tempurl}


\end{thebibliography}

\appendix

\end{document}